\documentclass[copyedit]{informs1}  
\pdfoutput=1

\usepackage{multirow}
\usepackage{endnotes}
\usepackage{enumerate}
\usepackage{rotating}
\usepackage{subfigure}
\usepackage{graphicx}
\usepackage{multirow}
\usepackage{hhline}
\usepackage{amsmath}
\usepackage{eufrak}

\usepackage{setspace,graphics,booktabs,color}
\usepackage{multirow}

\usepackage{float}
\floatstyle{ruled}
\newfloat{algorithm}{htp}{lop}
\floatname{algorithm}{Algorithm}

\usepackage{natbib}
 \bibpunct[, ]{(}{)}{,}{a}{}{,}%
 \def\bibfont{\small}%
\makeatletter

\newcommand{\Rmnum}[1]{\expandafter\@slowromancap\romannumeral #1@}
\makeatother

\newcommand{\mD}{\mathcal{D}}

\newcommand{\fb}{f_b}
\newcommand{\fl}{f_l}

\TheoremsNumberedThrough     

\EquationsNumberedThrough    

\definecolor{DSgray}{cmyk}{0,1,0,0}

\begin{document}

%
%
%
%
%
%
%
%
%
%
%

\RUNAUTHOR{}
%
\RUNTITLE{Model-Agnostic Linear Competitors}

\TITLE{\Large Model-Agnostic Linear Competitors - When Interpretable Models Compete and Collaborate with Black-box Models}

\ARTICLEAUTHORS{
	
	\AUTHOR {Hassan Rafique} \AFF{Program in Applied Mathematical and Computational Sciences (AMCS), University of Iowa, Iowa City, IA 52242, {hassan-rafique@uiowa.edu}}
	
	\AUTHOR {Tong Wang} \AFF{Tippie College of Business, University of
		Iowa, Iowa City, IA, 52245, {tong-wang@uiowa.edu}} 
	
	\AUTHOR {Qihang Lin} \AFF{Tippie College of Business, University of
		Iowa, Iowa City, IA, 52245, {qihang-lin@uiowa.edu}}
	
}

\ABSTRACT{Driven by an increasing need for model interpretability, interpretable models have become strong competitors for black-box models in many real applications. In this paper, we propose a novel type of model where interpretable models compete and collaborate with black-box models. We present the Model-Agnostic Linear Competitors (MALC) for partially interpretable classification. MALC is a hybrid model that uses linear models to locally substitute any black-box model, capturing subspaces that are most likely to be in a class while leaving the rest of the data to the black-box. MALC brings together the interpretable power of linear models and good predictive performance of a black-box model. We formulate the training of a MALC model as a convex optimization. The predictive accuracy and transparency (defined as the percentage of data captured by the linear models) balance through a carefully designed objective function and the optimization problem is solved with the accelerated proximal gradient method.  Experiments show that MALC can effectively trade prediction accuracy for transparency and provide an efficient frontier that spans the entire spectrum of transparency. }



\maketitle
%


\section{Introduction}
With the rapid growth of data in volume, variety and velocity \citep{zikopoulos2012understanding}, there has been increasing need for modern machine learning models to provide accurate and reliable predictions and assist humans in decision making. The interaction of models and humans naturally calls for users' understanding of machine learning models, especially in high-stake applications such as healthcare, judiciaries, etc \citep{letham2015interpretable,yang2018know,caruana2015intelligible,chen2018interpretable}. Thus, many state-of-the-art machine learning models such as neural networks and ensembles stumble in these domains  since they are \emph{black-box} in nature. Black-box models have an opaque or highly complicated decision-making process that is hard for human to understand and rationalize. Driven by the practical needs, researchers have shifted their focus from only predictive performance driven to also account for  transparency of models. It has recently been called by EU's General
Data Protection Regulation (GDPR)  for the  ``right to explanation'' (a right to information about individual decisions made by algorithms)  \citep{parliament,phillips2018international} that requires human understandable predictive processes of models.

The recent advances in machine learning has seen an increasing amount of interest and work in \emph{interpretable machine learning},  models and techniques that facilitate human understanding. Different forms of interpretable models have been developed, including rule-based models \citep{wang2017bayesian,lakkaraju2017interpretable}, scoring models \citep{zeng2017interpretable}, case-based models \citep{richter2016case}, etc. While these models can sometimes perform as well as black-box models, the performance loss is often inevitable, especially when the data is large and complex.  This is because black-box models are  optimized only for the predictive performance while interpretable models also pursue the small complexities. These two types of models have been competitors and mutally exclusive choices for users.

Another popular form of models have also risen quickly to assist human understandability, black-box explainers. Since the first paper of LIME \citep{ribeiro2016should}, a local linear explainer of any black-box model, various explainer models have been proposed \citep{ribeiro2018anchors,lundberg2017unified}. The main idea of explainers is they use simple and easily understandable models like decision rules or linear models, to locally or globally approximate the predictions of black-box models, providing ``post-hoc'' explanations with these simpler replica. However serious concerns have been brought up  \citep{rudin2019please, aivodji2019fairwashing,Thibault2019} on potential issues of black-box explainers since explainers only approximate but do not characterize exactly the decision-making process of a black-box model, often yielding an imperfect fidelity  to the original black-box model. In addition, there exists ambiguity and inconsistency \citep{ross2017right,lissack2016dealing} in the explanation since there could be different explanations for the same prediction generated by different explainers, or by the same explainer with different parameters. There's a very recent work that demonstrates that explanations can be deceptive and contrary
 to the real mechanism in a model \citep{aivodji2019fairwashing}. \citet{Alvarez2018} showed  LIME’s explanation of two close points (similar instances) can vary greatly. This is because LIME uses other instances in the neighborhood of the given instance to evaluate the local linear approximation. There are no clear guidelines for choosing an appropriate neighborhood that works the best and changing neighborhoods leads to a change in the explanations. This instability in the explanation demands a cautious and critical use of LIME. 
All of the issues result from the fact that the explainers only approximate in a post hoc way. They are not the decision-making process themselves.

In this paper, we propose a new form of model, which combines the intuitive power of interpretable models and the good predictive performance of  black-box models, to reach some controllable middle ground where both transparency and good predictive performance is possible.
The idea is simple and straightforward, a complex black-box model may have the best predictive performance overall, but it is not necessarily the best \emph{everywhere} in the data space. Some instances may be accurately predicted by simpler models instead of the black-box model without losing any (or intolerable) predictive performance. 

We design a unique mechanism where interpretable models complete and collaborate with a black-box model. Given a $K$ class classification problem, we design $K+1$ models, which we call \emph{agents}. $K$ of the agents are \emph{interpretable}, capturing $K$ classes, respectively. The remaining one is a pre-trained black-box model, called agent $\mathcal{B}$. Given an input $\mathbf{x}$, all of the $K$ agents bid to claim the input by proposing a score. The input is then assigned to the highest bidder with a significant margin over the other agents' scores. If there does not exist a winner (not winning by a large margin), then none of the $K$ agents can claim the input, and it is then sent to agent $\mathcal{B}$ by default. At agent $\mathcal{B}$, the input will be classified, and this classification process is unknown to other agents the whole time, i.e., \emph{model-agnostic}. 

\begin{figure}[h]
\centering
    \includegraphics[trim={0 0 0 0.8cm},clip, width=0.8\linewidth]{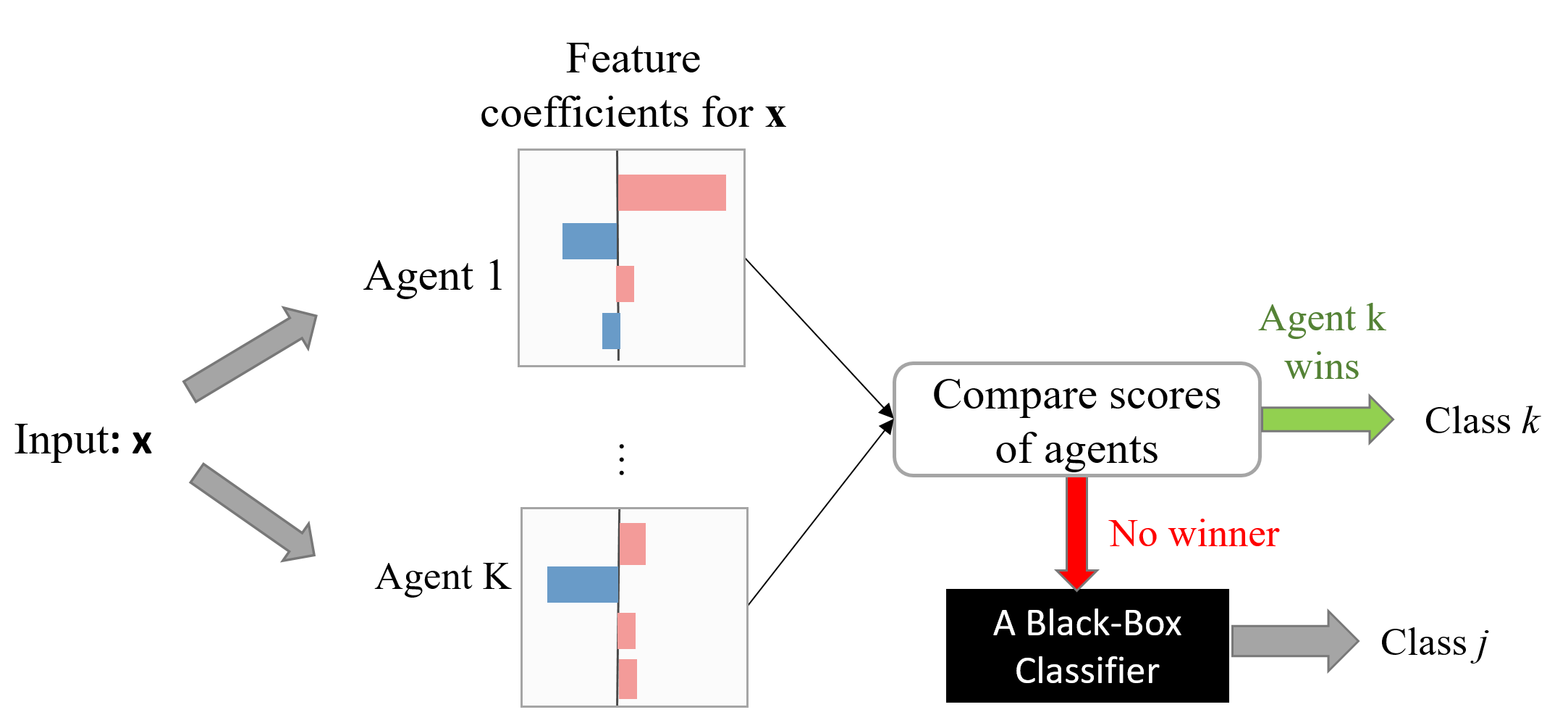}
\caption{The decision-making process of MALC.
}\label{fig:framework}
\end{figure}

In this paper, we let all interpretable agents be linear models, which is one of the most popular forms of interpretable models. The black-box model can be \emph{any} pre-trained multi-class classifier. We propose a model called Model-Agnostic Linear Competitors (MALC).
MALC partitions the feature space into $K+1$ regions, each claimed by an agent.  Agent $k (1 \leq k \leq K)$ captures the most representative and confident characteristics of class $k$ by claiming the most plausible area for class $k$. Predictions for this area are inherently interpretable since the agents are linear models with regularized numbers of non-zero coefficients.  The unclaimed area represents the subspace where none of the interpretable agents are very certain about, thus left to the most competent black-box agent $\mathcal{B}$. See Figure \ref{fig:framework} for an illustration. Meanwhile, the coefficients of the $K$ linear models also show the most distinctive characteristics of each class, providing an intuitive description of the classes.

To train MACL, we formulate a carefully designed convex optimization problem which considers the predictive performance, interpretability of the linear agents (coefficients regularization), and most importantly, the percentage of the area claimed by the linear model, which we define as \textbf{transparency} of MALC. Then we use accelerated proximal gradient method~\citep{nesterov2013introductory} to train MALC. 
By tuning the parameters, MACL can decide to send more or less area to the linear competitors, at the possible cost of the predictive performance. 

Our work is differentiated from linear explainers such as Local Interpretable Model-agnostic Explainers (LIME) \citep{ribeiro2016should}, which provide post hoc approximations or explanations but do not participate in the predictive performance. Here our linear models directly compete with the black-box model to generate predictions, equivalent to locally substituting the black-box on a subset of data. Thus, MALC avoids some of the controversial issues of black-box explainers such as ambiguity and inconsistency in the explanations. In addition, LIME provides a local explanation for an instance while MACL characterizes a more global description of each class since it captures subspaces of classes.

The rest of the paper is organized as follows. We review related work in Section \ref{sec:related}. The model is presented in Section \ref{sec:model}, where we formulate the model and describe the training algorithm. We conduct an experimental evaluation in Section \ref{sec:exp} on public datasets where MALC collaborates with state-of-the-art classifiers. 

\section{Related Work}\label{sec:related}
Our work is related but different from recent black-box explainers. MALC does not explain or approximate the behavior of a black-box model, but instead, collaborates with the black-box model and shares the prediction task.

We have found a few works in the literature on the combination of multiple models \citep{kohavi1996scaling,towell1994knowledge}. For example, \citep{kohavi1996scaling} combined a decision tree with a Naive Bayes model,  \citep{shin2000hybrid} proposed a system combining neural network and memory-based learning, \citep{hua2006hybrid} combined SVM and logistic regression, etc. A recent work \citep{wang2015trading} divides feature spaces into regions with sparse oblique tree splitting and assign local sparse additive experts to individual regions. Besides these more isolated efforts, there has been a large body of continuous work on neural-symbolic or neural-expert systems \citep{garcez2015neural} pursued by a relatively small research community over the last two decades and has yielded several significant results \citep{mcgarry1999hybrid,garcez2012neural,taha1999symbolic,towell1994knowledge}.   This line of research has been carried on to combine deep neural networks with expert systems to improve predictive performance \citep{hu2016harnessing}.

Compared to the models discussed above, our method is distinct in that it is model-agnostic and can work with \textbf{\emph{any}} black-box classifier. The black-box can be a carefully calibrated, advanced model using confidential features or techniques. Our model only needs predictions from the black-box and does not need to alter the black-box during training or know any other information from it. This minimal requirement of information from the black-box collaborator renders much more flexibility in creating collaboration between different models, largely preserving confidential information from the more advanced partner.

One work that's closest to ours is Hybrid Rule Sets (HyRS) \citep{wanghybrid2019} that builds a hybrid of decision rules and a black-box model. An input goes through a positive rule set, a negative rule set, and a black-box model sequentially until it is classified by the first model that captures it. HyRS produce interpretable predictions on instances captured by rules. A HyRS only works with binary classification.   MALC, on the other hand, can work with multi-class classification.  The $K$ interpretable agents compete for an input simultaneously in a fair mechanism.

\section{Model-Agnostic Linear Competitors}\label{sec:model}

\begin{figure}[h]
 \centering
   \includegraphics[width=0.32\linewidth]{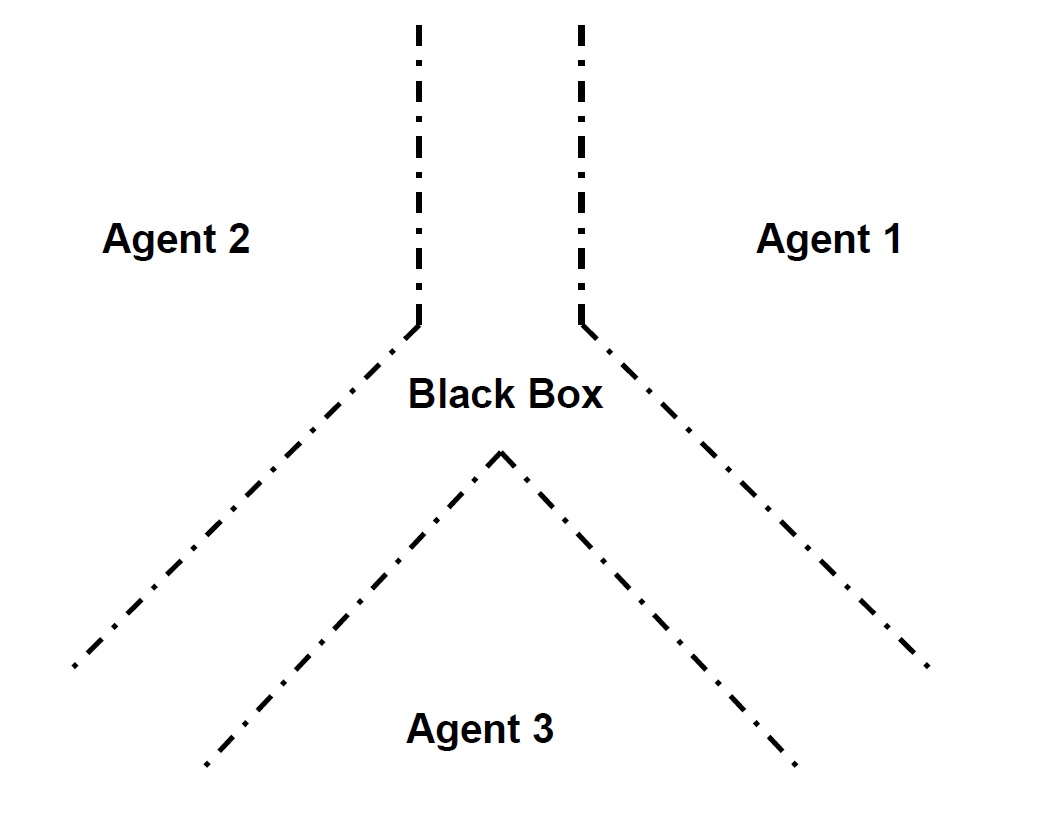}
 \caption{A simplified depiction of partitioning of the data space in the case of three classes, by linear and black-box agents. }\label{fig:dataspace}
 \end{figure}

In this paper, we focus on the multi-class classification problem. Suppose there are $K$ distinct classes. We consider an approach similar to one-vs-all linear classification. We us review how this classification works. Given a linear classifier $f_i(\mathbf{x}) = \mathbf{w}^\top_i\mathbf{x}$,  $i \in [K]:= \{1,2,...,K\}$, if $f_i(\mathbf{x}) - f_j(\mathbf{x}) \geq 0$, for every $j$ other than $i$, then $\mathbf{x}$ belongs to class $i$.
For class $i$, \[\mathcal{P}_i(\mathbf{x}) = \bigcap_{j\not = i} \bigg\{ f_i(\mathbf{x}) - f_j(\mathbf{x}) = 0 \bigg\} \]
is the decision boundary. Most mistakes made by a linear model happen around the decision boundary. Therefore, in a hybrid model, we exploit the high predictive power of a black-box model and leave this more difficult area to it while having the linear classifier classify the rest. Then the linear classifier produces a decision only when it is confident enough, this time comparing against thresholds $\{\theta_i \geq 0\}^K_{i=1}$: to predict class $i$ when $f_i(\mathbf{x}) - f_j(\mathbf{x}) \geq \theta_i$ for every $j$ other than $i$ and unclassified otherwise. Thus the linear model generates $K$ decision boundaries, creating a partition of a data space into $K+1$ regions, a region for each of the $K$ classes and an unclassified region. This unclassified region contains data that the linear model is not confident to decide so that black-box is activated to generate predictions., see Figure \ref{fig:dataspace}. 
Thus we build $K$ linear competitors, each advocating for a class, to collaborate with the black-box model. We call this classification method Model-Agnostic Linear Competitors (MALC) model.

The goal of building such a collaborative linear model is to replace the black-box system with a transparent system on a subset of data, at the minimum loss of predictive accuracy. Therefore a key determinant in the success of MALC is the partitioning of the data, which is determined by the coefficients $\mathbf{w}_i$ in the linear model and the thresholds $\theta_i$, $i \in [K]$.
 In this paper, we formulate a convex optimization problem to learn the coefficients and thresholds. The objective function considers the fitness to the training data, captured by a convex loss function, the regularization term, and the sum of thresholds. As $\theta_i$ gets close to $0$, more data can be decided by the linear model, increasing the transparency of the decision-making process, but at the cost of possible loss of predictive performance. Our formulation is compatible with various forms of convex loss function and guarantees global optimality. 

We work with a set of training examples $\mathcal{D} = \{(\mathbf{x}_i,y_i)\}_{i=1}^n$ where $\mathbf{x}_i \in \mathbb{R}^d$ is a vector of $d$ attributes and $y_i\in [K] := \{1,2,...,K\}$ is the corresponding class label. 
Let $f(\mathbf{x}):\mathbb{R}^d\rightarrow [K]$ represent the MALC classification model that is constructed based on linear models $f_{l,i}(\mathbf{x})=\mathbf{w}^\top_i\mathbf{x}$, $i\in [K]$ and a black-box model $f_b(\mathbf{x}):\mathbb{R}^d\rightarrow[K]$. The black-box model is given, which can be \emph{any} trained model. We need its prediction on the training data  $\mathcal{D}$, denoted as $\{y^b_i\}_{i=1}^n$ and $y^b_i=f_b(\mathbf{x}_i)$. Our goal is to learn the coefficients $\mathbf{w_i}$ in the linear models $f_{l,i}$ together with  thresholds $\theta_i$ ( $\geq 0$), $i \in [K]$, in order to form a hybrid decision model $f$ as:
\begin{eqnarray}
\label{eq:hybridf}
f(\mathbf{x})=
\left\{
\begin{array}{cl}
     k &  \text{ if } \mathbf{w}_k^\top\mathbf{x} - \mathbf{w}_j^\top\mathbf{x}  \geq \theta_k,~ k \in [K],~\forall ~ j \in [K]\setminus \{k\}\\
     f_b(\mathbf{x})& ~\text {otherwise}
\end{array}
\right.
\end{eqnarray}

Note the hybrid model uses $K$ thresholds to partition the data space into $K+1$ regions, a region for each class and an undetermined region left to the black-box model. Data that falls into any of the $K$ class's claimed regions is considered ``transparent'' by the linear model, and we refer to the percentage of this data subset as the \textbf{transparency} of the model. 

\subsection{Model Formulation}
In this section, we formulate an optimization framework to build a MALC model.
We consider three factors when building the model: \emph{predictive performance}, \emph{data transparency}, and \emph{model regularization}. We elaborate each of them below.

The (in-sample) predictive performance characterizes the fitness of the model to the training data. Since $\fb$ is pre-given, the predictive performance is determined by two factors, the accuracy of $\fl = [f_{l,1},~f_{l,2},...~,f_{l,K}]$ on instances as described in ($\ref{eq:hybridf}$) and the accuracy of $\fb$ on the remaining examples. We wish to obtain a good partition of data $\mD$ by assigning $\fb$ and $\fl$ to a different region of the data such that the strength of $\fb$ and $\fl$ are properly exploited. Second, we include the sum $\sum \theta_i$ as a penalty term in the objective to account for data transparency of the hybrid model. The smaller sum implies more data is classified by the linear model. In the most extreme case where $\sum \theta_i = 0$, all data is sent to the linear model, and the MALC model is reduced to a pure one-vs all linear classifier, i.e., transparency equals one. 
Finally, we also need to consider model regularization in the objective. As the weight for the sparsity enforcing regularization term increases, the model encourages using a smaller number of features which increases the interpretability of the model as well as preventing overfitting.

Combining the three factors discussed above, we formulate the learning objective for MALC as: 
\begin{equation}
\label{eq:obj}
F^*:=\min_{\mathbf{w},\theta\geq 0}\left\{F(\mathbf{w},\theta):=\mathcal{L}(\mathbf{w},\theta;\mD) + C_1 \sum^K_{i=1}\theta_i + C_2 r(\mathbf{w})\right\}, 
\end{equation}
where $\mathbf{w} = [\mathbf{w}_1,~\mathbf{w}_2,...~,\mathbf{w}_K]$, $\mathbf{\theta} = [\theta_i,~\theta_2,...~,\theta_K]$, $\mathcal{L}(\mathbf{w},\theta;\mD)$ is the loss function defined on the training set $\mD$ associated to the decision rule $f$ in~\eqref{eq:hybridf}, $\sum^K_{i=1}\theta_i$ is a penalty term to increase the transparency of $f$, $r$ is a convex and closed regularization term (e.g. $\|\mathbf{w}\|_1$, $\frac{1}{2}\|\mathbf{w}\|_2^2$ or an indicator function of a constraint set), and $C_1$ and $C_2$ are non-negative coefficients which balance the importance of the three components in~\eqref{eq:obj}. 

Let $I_k = \{i~|~ y_i = k\}$, which is the index set of all the data points $(\mathbf{x})$ belonging to class $k$. Similarly, let $I^+_{k} = \{i \in I_k~|~ y^i_b = y_i\}$ and $I^-_{k} = \{i \in I_k~|~ y^i_b \not = y_i\}$.  The loss function in \eqref{eq:obj} over the dataset $\mathcal{D}$ is then defined as 
\begin{equation}
\label{eq:loss}
\mathcal{L}(\mathbf{w},\mathbf{\theta};\mD) = 
\frac{1}{n}\sum^K_{k = 1} \sum_{i\in I^+_k} \sum^K_{j  = 1 \atop j \not = k} 
\phi(\mathbf{w}_k^\top \mathbf{x}_i - \mathbf{w}_j^\top \mathbf{x}_i + \theta_j ) +
\frac{1}{n} \sum^K_{k = 1} \sum_{i\in I^-_k} \sum^K_{j  = 1 \atop j \not = k} 
\phi(\mathbf{w}_k^\top \mathbf{x}_i - \mathbf{w}_j^\top \mathbf{x}_i - \theta_k )  
\end{equation}

where function $\phi(z):\mathbb{R}\rightarrow \mathbb{R}$ is a non-increasing convex closed loss function which can be one of those commonly used in linear classification such as the hinge loss $\phi(z)=(1-z)_+$, smooth hinge loss $\phi(z)= \frac{1}{2}(1-z)^2_+$ or the logistic loss $\phi(z)=\log(1+\exp(-z))$. Note that $\{I_k = I^+_{k} \cup I^-_{k}\}^K_{k=1}$ form a partition of $\{1,2,\dots,n\}$. The intuition of this loss function is as follows. Take a data point $\mathbf{x}_i$ with $y_i = k$ and $y^b_i = k$ as an example. Our hybrid model \eqref{eq:hybridf} will classify $\mathbf{x}_i$ correctly as long as it does not fall into the region of a class other than $k$. To ensure $\mathbf{x}_i$ does not fall into another class's region, we need  $\mathbf{w}^\top_j \mathbf{x}_i - \mathbf{w}^\top_k \mathbf{x}_i  < \theta_j$ for every $j$ other than $k$. Hence, with the non-increasing property of $\phi$, the loss term $\phi(\mathbf{w}_k^\top \mathbf{x}_i - \mathbf{w}_j^\top \mathbf{x}_i + \theta_j )$ will encourage a positive value of $\mathbf{w}_k^\top \mathbf{x}_i - \mathbf{w}_j^\top \mathbf{x}_i + \theta_j $ which means we have $\mathbf{w}_j^\top \mathbf{x}_i - \mathbf{w}_k^\top \mathbf{x}_i < \theta_j$. On the other hand, for a data point $\mathbf{x}_i$ with $y_i = k$ and $y^b_i \not = k$, our hybrid model will classify $\mathbf{x}_i$ correctly only when $\mathbf{x}_i$ falls in the class $k$ region, namely, $\mathbf{w}^\top_k \mathbf{x}_i - \mathbf{w}^\top_j \mathbf{x}_i \geq \theta_k$ for every $j$ other than $k$. Hence, we use the loss term $\phi(\mathbf{w}_k^\top \mathbf{x}_i - \mathbf{w}_j^\top \mathbf{x}_i - \theta_k )$ to encourage a positive value of $\mathbf{w}_k^\top \mathbf{x}_i - \mathbf{w}_j^\top \mathbf{x}_i - \theta_k $.

\subsection{Model Training}

With the loss function defined in \eqref{eq:loss}, the hybrid model can be trained by solving the convex minimization problem~\eqref{eq:obj} for which many efficient optimization techniques are available in literature including subgradient methods~\citep{nemirovski2009robust,duchi2011adaptive}, accelerated gradient methods~\citep{nesterov2013introductory,beck2009fast}, primal-dual methods~\citep{nemirovski2004prox,chambolle2011first} and many stochastic first-order methods based on randomly sampling over coordinates or data~\citep{johnson2013accelerating,duchi2011adaptive}. 
The choice of algorithms for~\eqref{eq:obj} depends on various characteristics of the problem such as smoothness, strong convexity, and data size. 

Since numerical optimization is not the focus of this paper, we will simply utilize the accelerated proximal gradient method (APG) by Nesterov~\citep{nesterov2013introductory} to solve \eqref{eq:obj} when $\phi$ is smooth. 

\section{Experiments}\label{sec:exp}
We perform a detailed experimental evaluation of the proposed model on four public datasets. The goal here is to examine the predictive performance, the transparency, and the model complexity. In addition, we characterize the trade-off between predictive accuracy and transparency using \textbf{efficient frontiers}. To do that, we vary the parameters $C_1$ and $C_2$ to generate a list of models producing an accuracy-transparency curve for each dataset.  We also analyze a medical dataset in detail to provide users more intuitive understanding of the model.

\paragraph{Datasets} We analyze four real-world datasets that are publicly available at \citep{LIBSVM, Kaggle,quinlan1986inductive, wang2017bayesian}. 
  1) \emph{Coupon} \citep{wang2017bayesian} (12079 $\times$  113) studies responses of consumers to recommendation of coupons   when users are driving in different contexts, using feature such as the passenger, destination, weather, time, etc. The three classes are ``decline'', ''accept and will use right away'', and ``accept and will use later''  2) \emph{Covtype}\citep{LIBSVM} (581,012 $\times$ 54) studies the forest cover type of wilderness areas which include Roosevelt National Forest of northern Colorado. There are seven different forest cover types. The features in the covtype dataset are scaled to $[0,1]$. 3) \emph{Thyroid}\citep{quinlan1986inductive} (9172 $\times$  63)  studies the prediction of thyroid diagnoses based on patients' biomedical information. 4) \emph{Medical} \citep{Kaggle} (106,643 $\times$ 14) provide information about Clinical, Anthropometric and Biochemical (CAB) survey done by Govt. of India. This survey was conducted in nine states of India with a high rate of maternal and infant death rates in the country. We focused on the subset of data for children under the age of five and predicted their illness type. We dropped some features not needed for classification, and the missing values in certain features were replaced by mean or mode values appropriately. 
  For each dataset, we randomly sample 80\% instances to form the training sets and use the remaining 20\% as the testing sets. Since the \emph{Medical} dataset is highly unbalanced among different classes, we  downsample the majority class and upsample the minority class to make them balanced.

  \paragraph{Training Black-box Models} We first choose three state-of-the-art black-box classifiers, Random Forest \citep{liaw2002classification}, XGBoost \citep{chen2016xgboost} and fully-connected neural network with two hidden layers. All of these models are implemented with R. The Random Forest model is built using the \emph{ranger} package~\citep{wright2015ranger}.  The XGBoost model is built using the \emph{xgboost} package~\citep{chen2015xgboost}. The neural network model is built using the \emph{keras} package~\citep{chollet2017kerasR}.  For each model, we identify one or two hyperparameters and, for each dataset, we apply an $80\%$-$20\%$ holdout method on the training set to select the values for these hyperparameters from a discrete set of candidates that give the best validation performance. For Random Forest, we use $500$ trees and tune the minimum node size and maximal tree depth. For XGBoost, we tune maximal tree depth and the number of boosting iterations. For the neural network, we choose the sigmoid function to be the activation function and tune the number of neurons and the dropout rates in the two hidden layers.

    \paragraph{Training MALC} We use the predictions of the three black-box models on the training set as the input to build MALC models. In \eqref{eq:obj}, we choose $\phi$ to be the smooth hinge loss and $ r(\mathbf{w}) = \|\mathbf{w}\|_1$. We would like to obtain a list of models that span the entire spectrum of transparency, so we vary $C_1$ and $C_2$ to achieve that goal. Note that $C_1$ is directly related to transparency, and we use grid-search to find a suitable range to achieve transparency from zero to one. $C_2$ is related to the sparsity of the model. Overall, we choose $C_1$ from $[0.005, 0.95]$ and $C_2$ from $[0.03, 0.25]$. For each $C_1$ value, we use $80\%$-$20\%$ holdout on the training set to choose $C_2$ from a discrete set of candidates that give the best validation performance. After choosing the pairs of $(C_1,C_2)$ values, the Algorithm APG is run up to $10,000$ iterations to make sure the change in objective value was less than $0.1 \%$, in the last iterations, to ensure the convergence.   

  \begin{figure}[hbt!]
    \centering
    \includegraphics[scale=0.1975]{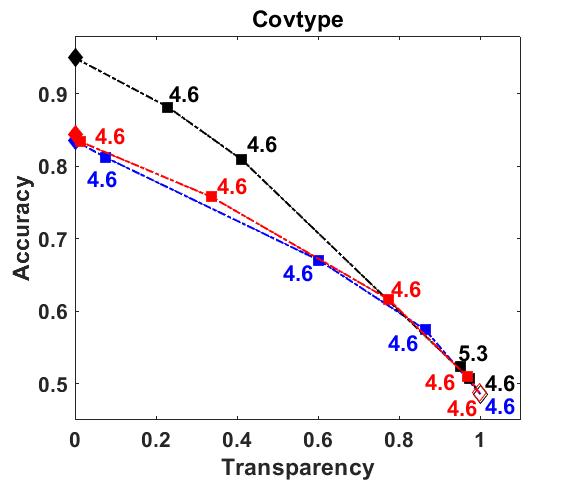}
    \includegraphics[scale=0.1975]{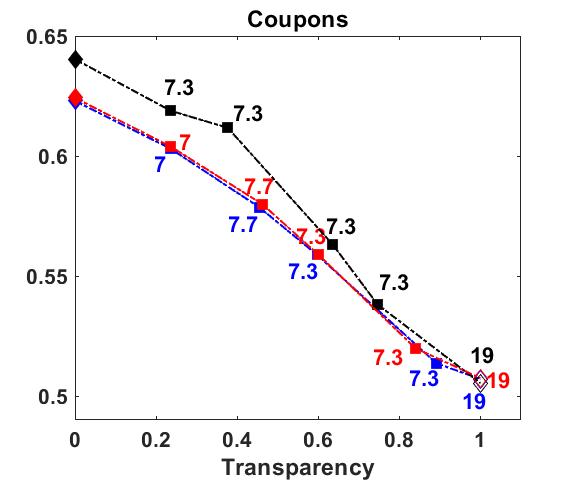}
    \includegraphics[scale=0.1975]{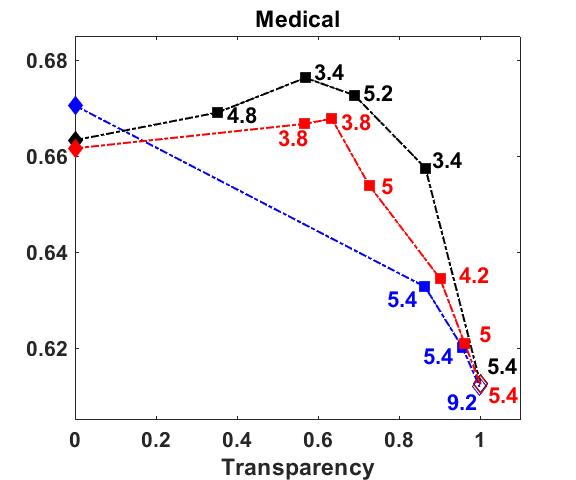}
    \includegraphics[scale=0.1975]{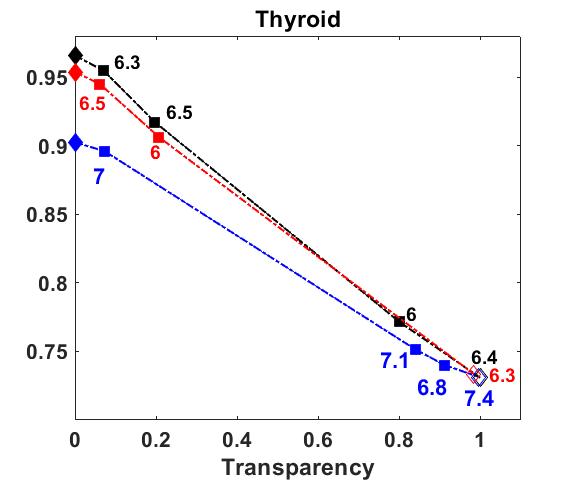}\\
    \includegraphics[scale=0.45]{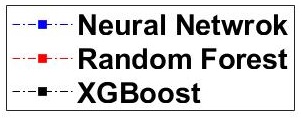}
    \caption{ The efficient frontiers of MALC when collaborating with different black-box models. The numbers represent the average number of features being used by the MALC model.}
    \label{fig:EF-Curves}
\end{figure}

\paragraph{Efficient Frontier Analysis}
In Figure \ref{fig:EF-Curves}, each efficient frontier starts with a transparency value of zero, which corresponds to a pure black-box model. The general trend is as transparency increases, accuracy tends to decrease. The medical dataset provides an interesting scenario where the initial increase in transparency does not lead to a decrease in predictive performance. The rate of change of transparency w.r.t predictive performance is different for each dataset. For Thyroid dataset, the accuracy decreases almost linearly, whereas, for coupon and covtype datasets, accuracy decreases steadily as the transparency increases. However, for the medical dataset, the accuracy does not decrease initially and then falls significantly after a certain transparency threshold. Note that the transparency value of one corresponds to a pure linear (interpretable) model. But the interpretability comes at a huge cost of predictive performance, as evident by considerably low accuracy of linear models compared to the accuracy of the black-box models for all datasets. 
MALC provides the user with a unique framework of choosing a model from the whole spectrum of options available on an efficient frontier with their desired accuracy and transparency. We recommend the users to choose the models around the tipping point to ensure gain in transparency without a significant loss in accuracy.
\paragraph{Number of Features Analysis} We would also like to make sure the linear models are indeed interpretable, i.e., using a few non-zero terms in the model. We report in Figure \ref{fig:EF-Curves} the average number of non-zero coefficients in MALC, which is calculated as a ratio of the number of non-zero coefficients in $K$ linear models ($\mathbf {w}$) to the number of classes in the dataset.  Observe that MALC models require a relatively small number of features from the dataset to gain transparency, preserving the interpretability of linear models.

The control over transparency-accuracy trade-off and use of a small number of features to gain transparency make MALC a strong candidate for real-world applications, particularly when the user wants to avoid the black-box methods. 

\subsection{Case Study on the Medical Dataset}

\begin{figure}[hbt!]
    \includegraphics[width=1\textwidth]{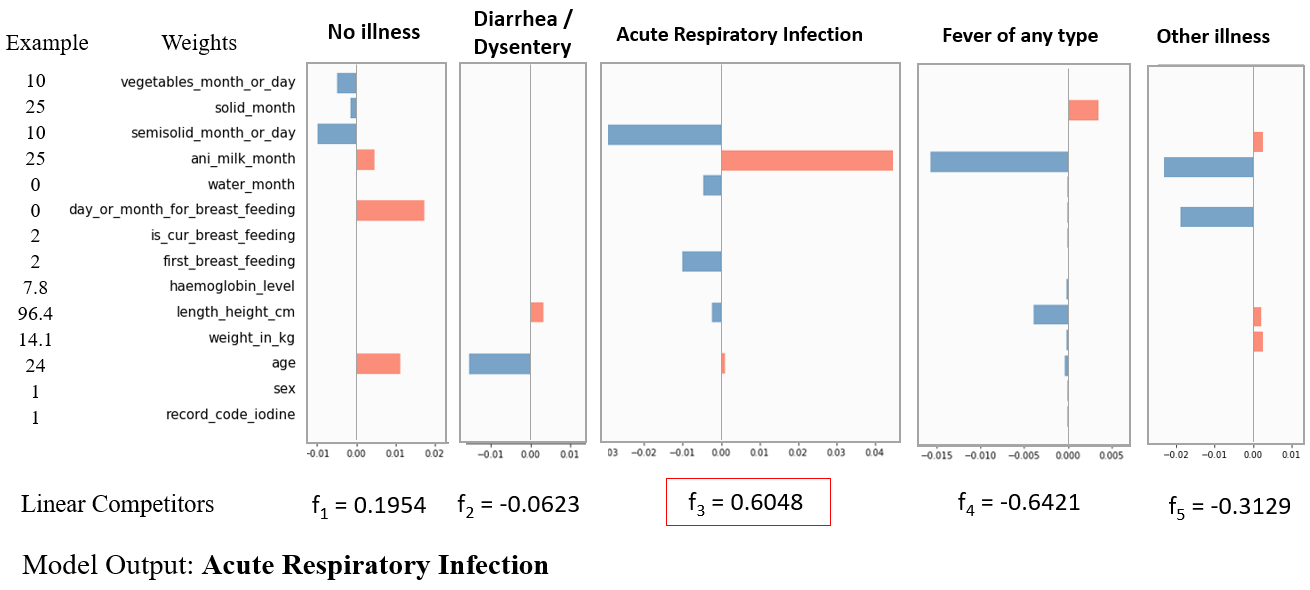}
    \caption{ An example of MALC in collaboration with a pre-trained random forest}
    \label{fig:example}
\end{figure}

We show an example of MALC on the medical dataset. There are a total of five classes in this datsaet, ``no illness'', diarrhea/dysentery'', ``acute respiratory infection'', ``fever of any type'', and ``other illness''. MALC was built in collaboration with a pre-trained random forest whose accuracy is 66.0\%. After building five linear competitors, the accuracy of MALC reaches 66.4\% while gaining transparency of 77.7\%. The coefficients of  the five linear models are shown in Figure \ref{fig:example}.  From the linear models, one can easily extract some of the key characteristics for each class. For example, the later children start to receive semisolid food and the longer they are exclusively breastfed (feature ``day\_or\_month\_for\_breast\_feeding )'', the more likely they will be free of any of the illness (Class 1). Children who start receiving semisolid mashed food (feature ``semisolid\_month\_or\_day'') at a very young age, start receiving water at an early age (feature ``water\_month''), and are too late to start receiving animal milk/formula milk (feature ``ani\_milk\_month'') are more likely to have acute respiratory infection (Class 3).

We chose an example instance and show the input features and the output of the linear models in Figure \ref{fig:example}. This child started receiving animal milk/formula milk at age of $25$ months, almost six times of the average age of receiving animal/formula milk (4.3 months). 
This child started receiving semisolid food at 10 months old, later than the average age of children (5.8 months) who start receiving semisolid food. This is helpful for the child's overall health conditions as suggested by classifier 1. However, this effect is completely overtaken by the late usage of formula milk. 

In addition, the child was breast fed later than $64\%$ of the children in the dataset. Combining these important features, classifier 3 outputs the highest score, with a large enough margin over the other four linear models. Thus this child is predicted to have acute respiratory infection, which is consistent with the true label.

An interesting observation for this model is it performs slightly better than the black-box alone, which means the 77.7\% transparency is obtained for free. This is the desired situation for hybrid models like MALC to be adopted.

\subsection{Comparison with baselines}

There are two lines of work in interpretable machine learning, stand-alone interpretable models like decision trees and black-box explainers like LIME. MACL has a unique model form does not fall into either of them. We choose representative models from each category. We compare with three decision trees as stand-alone interpretable models and LIME as an explainer. We focus and present results on the medical data. First, comparison with decision trees show that interpretable models are insufficiant for this dataset since they generate lower accuracy. In addition, we report the size of trees represented by the number of nodes in a tree to quantify the model complexity. Decision trees have significantly larger model sizes, as reported in Table \ref{tab:tree}.
\vspace{0.15in}
\begin{table}[h]
\small
\centering
\vspace{-4mm}
\caption{Performance comparison with baseline models}\label{tab:tree}
\begin{tabular}{l|ccclll}\toprule
        & MALC   & CART   & C4.5  & C5.0  \\ \hline
accuracy      &  0.66                                   &  0.63      &   0.63       &  0.62           \\ 
\# of rules      & --                                  &    84    &    46      &    67          \\
\# of conditions &  26 non-zero coefficients from 5 linear models                                     &  167      &   91       &   132            \\ \bottomrule
\end{tabular}
\end{table}\normalsize

\begin{figure}[hbt!]
    \vspace{-2mm}
    \includegraphics[width =\textwidth]{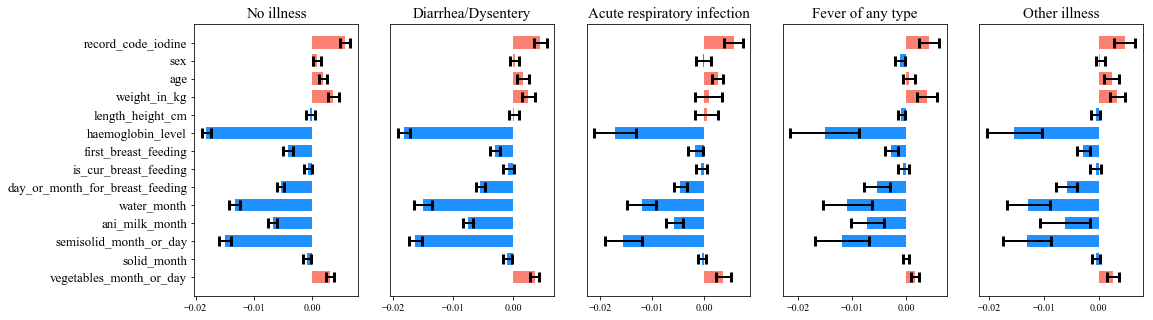}
    \caption{ Means and Standard Deviations of Coefficients of LIME}
    \label{fig:LIME}
\end{figure}

For comparison with LIME, we sample up to $200$ examples from each of the five classes that are explained by one of the linear classifiers in MALC and use LIME to generate explanations for each of them. We observe two issues with LIME. First, the inter-class explanations of LIME are too similar, as shown by the mean and std of the coefficients of LIME in Figure \ref{fig:LIME}: the means of  are almost identical across classes. This makes it difficult for users to understand the difference between classes and it's hard to use the explanations to reason why an instance is classified into a particular class but not others. Unlike LIME, MALC provides different coefficients for different classes (see Figure \ref{fig:example}) so that users can easily understand what features differentiate one class from the others.

Second, \citet{Alvarez2018} showed that LIME's explanation of two close (similar) instances can vary greatly. This is because LIME uses other instances in the neighborhood of the given instance to evaluate the local linear approximation. There are no clear guidelines for choosing an appropriate neighborhood that works the best and changing neighborhoods leads to a change in the explanations.  This instability in the explanation demands a cautious and critical use of LIME. On the other hand, MALC provides a set of \textit{global} linear models and is relatively independent of the local neighborhood. This means that explanations provided by MALC are more consistent and stable.

\section{Conclusion}
We proposed a Model-Agnostic Linear Competitors (MALC) model for multi-class classification. MALC builds $K$ linear models to collaborate with a pre-trained black-box model. The data space is partitioned by MALC, into regions classified by the linear model and the black-box with linear decision boundaries. We formulated the training of a MALC model as convex optimization, where predictive accuracy and transparency balance through objective function. The optimization problem is solved with the accelerated proximal gradient method.

MALC is model-agnostic, which makes it flexible to collaborate with any black-box model, needing only their predictions on the dataset. In this paper, MALC collaborated with Random Forest, XGBoost, and Neural Networks to solve multiclass classification problems. Experiments show that MALC was able to yield models with different transparency and accuracy values by varying the parameters, thus providing more model options to users. In real applications, users can decide the operating point based on the efficient frontier. The decision will depend on knowing how much loss in accuracy is tolerable and how much transparency is desired in their application.

Compared to post hoc black-box explainers such as LIME, the linear models in MALC are predictive models, which guarantee 100\% fidelity on data that are claimed by them. Also, unlike linear explainers that provide local explanations, the linear models in MALC capture global characteristics of classes by building $K$ linear models at the same time to compete with each other. Thus the coefficients learned are often the most distinguishing features.

The proposed work offers a new perspective in building handshakes between interpretable and black-box models, in addition to using the former as the post hoc analysis to the latter in the current literature. Here we propose to build collaboration between the two to exploit the strength of both. Despite the difference in the goal, existing black-box model explainers such as LIME can still be applied to explain the subset of data sent to the black-box.

\bibliographystyle{ormsv080}
\bibliography{DRL1}

\newpage
\APPENDICES

\end{document}